\documentclass{article}
\usepackage{amsmath,epsfig}
\usepackage[preprint]{spconfa4}
\usepackage{xcolor}
\usepackage{cite}
\usepackage{booktabs}
\usepackage{arydshln}
\usepackage{multirow,subfigure}
\usepackage{amssymb}
\usepackage{bbding}
\usepackage{color}
\usepackage{bbding}
\usepackage{pifont}

\let\OLDthebibliography\thebibliography
\renewcommand\thebibliography[1]{
  \OLDthebibliography{#1}
  \setlength{\parskip}{0pt}
  \setlength{\itemsep}{0pt plus 0.3ex}
}

\begin{document}\sloppy

\def\x{{\mathbf x}}
\def\L{{\cal L}}

\title{A No-Reference Quality Assessment Method for Digital Human Head}
%
\name{Yingjie Zhou$^{1,2,3}$, Zicheng Zhang$^{1,3}$,Wei Sun$^{1,3}$ ,Xiongkuo Min$^{1,3}$,Xianghe Ma$^{1,3}$ and Guangtao Zhai$^{1,3,4}$}
\address{$^{1}$Institute of Image Communication and Network Engineering, Shanghai Jiao Tong University, China\\
$^{2}$School of Information and Control Engineering, China University of Mining Technology, China\\
$^{3}$Peng Cheng Laboratory, China\\
$^{4}$MoE Key Lab of Artificial Intelligence, AI Institute, Shanghai Jiao Tong University, China\\
zyj2000@cumt.edu.cn}

\maketitle

%

\begin{abstract}
In recent years, digital humans have been widely applied in augmented/virtual reality (A/VR), where viewers are allowed to freely observe and interact with the volumetric content. However, the digital humans may be degraded with various distortions during the procedure of generation and transmission. Moreover, little effort has been put into the perceptual quality assessment of digital humans. Therefore, it is urgent to carry out objective quality assessment methods to tackle the challenge of digital human quality assessment (DHQA). In this paper, we develop a novel no-reference (NR) method based on Transformer to deal with DHQA in a multi-task manner. Specifically, the front 2D projections of the digital humans are rendered as inputs and the vision transformer (ViT) is employed for the feature extraction. Then we design a multi-task module to jointly classify the distortion types and predict the perceptual quality levels of digital humans. The experimental results show that the proposed method well correlates with the subjective ratings and outperforms the state-of-the-art quality assessment methods.
\end{abstract}
\vspace{0.2cm}
\section{Introduction}
\label{sec:intro}

\noindent Digital humans are virtual characters with human images, human characters, and behavioral characteristics simulated by computers, which are regarded as the entrance to the metaverse. Thanks to the rapid development of computer graphics, digital humans have been gradually integrated into people's lives in many fields such as film industry, social media, tourism, etc. 
\let\thefootnote\relax\footnotetext{This work was supported in part by NSFC (No.62225112, No.61831015), the Fundamental Research Funds for the Central Universities, National Key R\&D Program of China 2021YFE0206700, Shanghai Municipal Science and Technology Major Project (2021SHZDZX0102), and STCSM 22DZ2229005. }
\begin{figure}[t]
    \centering
    \includegraphics[width = 8.7cm]{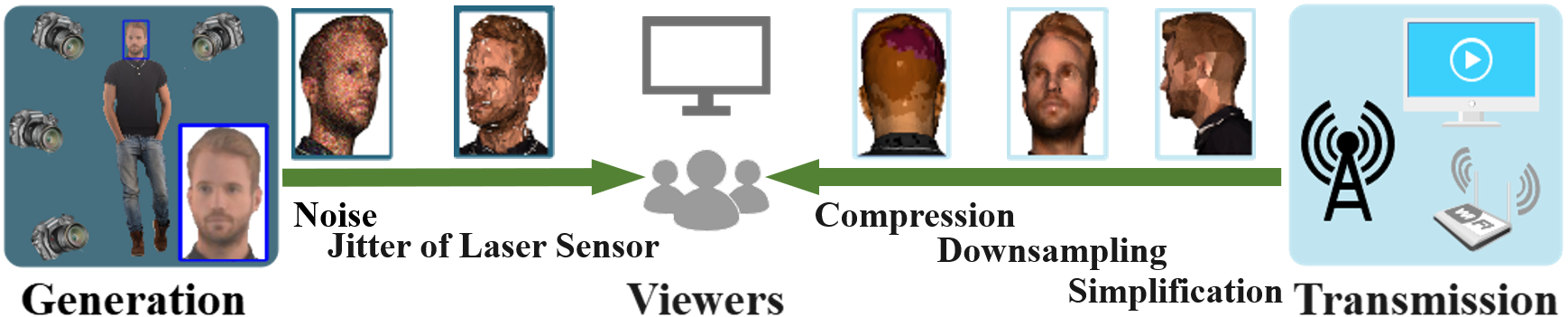}
    \vspace{-0.8cm}
    \caption{The generation of distortions. During the generation process of the digital humans, the jitter of the laser sensor and the thermal noise of devices can introduce geometry shift and noise to the digital human models. To meet the real-time needs of practical application scenarios where the transmission bandwidth is usually limited, the models are inevitably processed with compression, downsampling or simplification.}
    \label{fig:map}
    \vspace{-0.65cm}
\end{figure}
Unfortunately, as shown in Fig.~\ref{fig:map}, because the digital human models presented to viewers are often stored as textured meshes, whose geometry and color information are separately saved, many procedures may lead to the generation of distortions, which may severely damage the visual quality of digital humans. Therefore, it is important to carry out objective DHQA methods to optimize the transmission system and improve the quality of experience (QoE).\\
\begin{figure*}
\includegraphics[width=\textwidth]{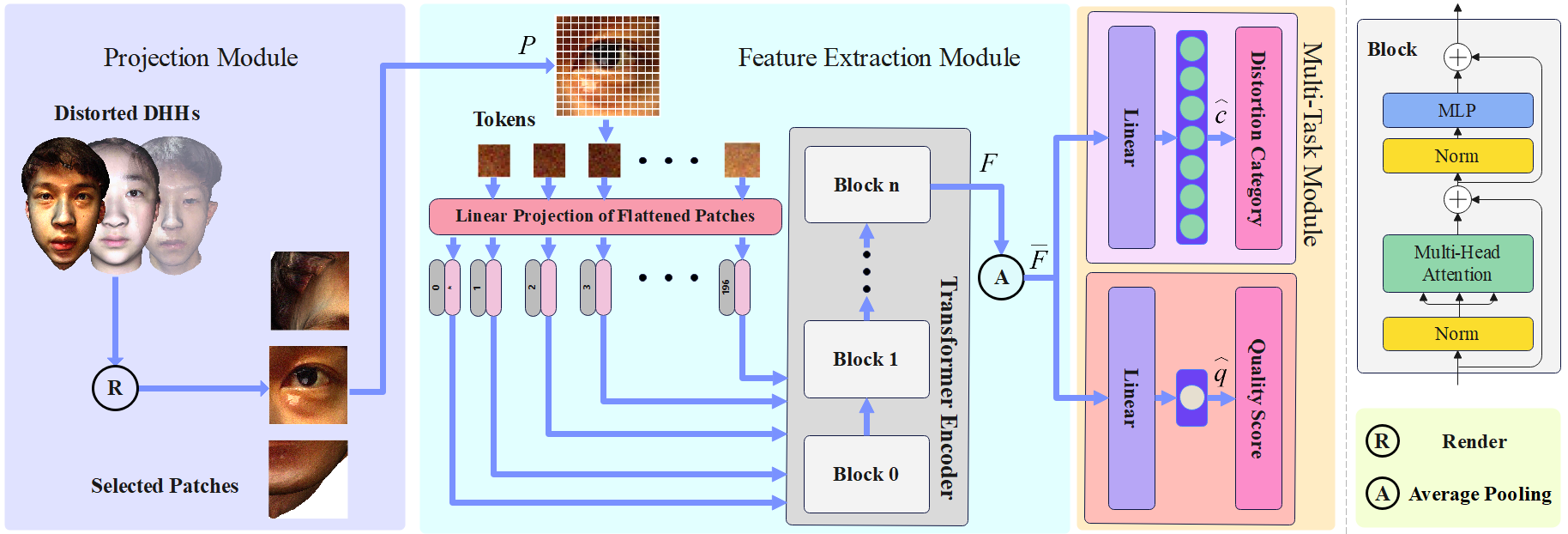}
\vspace{-0.8cm}
\caption{The framework of the proposed NR DHHQA model.} \label{fig1}
\vspace{-0.5cm}
\end{figure*}
\begin{figure}
    \centering
    \includegraphics[width = 8.1cm]{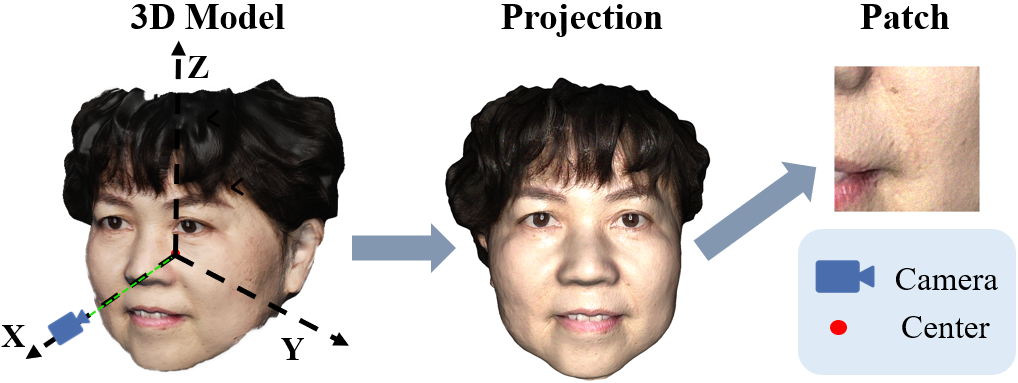}
    \vspace{-0.5cm}
    \caption{Illustration of the projection process.}
    \label{fig:projection}
    \vspace{-0.6cm}
\end{figure}
\indent Reviewing the development of 3D quality assessment (3DQA), many quality assessment metrics \cite{liu2022perceptual,nehme2020visual,liu2022point,yang2020predicting,nehme2022textured,zhang2022mm,zhang2022treating,zhang2022no} as well as 3DQA databases \cite{liu2022perceptual,yang2020predicting,nehme2020visual,zhang2022ddh} have been proposed. For 3DQA, the mainstream quality assessment methods can be divided into two categories: model-based methods\cite{liu2022perceptual,nehme2020visual,liu2022point} and projection-based methods\cite{yang2020predicting,nehme2022textured}. Specifically, the model-based method gives a quality score by calculating the distance between discrete points as a similarity measure. The projection-based method, on the other hand, projects the 3D model from different viewpoints and obtains the quality score of the model by evaluating the quality of the projection. \\
\indent In this paper, we mainly focus on the evaluation of digital human head (DHH), which is the most concerned part of digital humans.  In most practical occasions, DHHs are essential for representing facial features and expressions,  the perceptual quality of which directly affects the perceptual quality of digital humans. In addition, NR method is more valuable for practical applications and its research is more challenging because in many practical situations, we do not have access to the original reference model. Therefore, we propose a Transformer-based NR digital human head quality assessment (DHHQA) method in a multi-task manner. The proposed NR DHHQA method includes three modules: the projection module, feature extraction module and multi-task module. First, the projection module renders the front-side projections of the DHHs as inputs. Then, we select ViT\cite{dosoVITskiy2020image} as the feature extraction backbone and realize the effective extraction of quality-aware features. Finally, we get the quality-aware embedding by average pooling. In the design of the multi-task module, we divide the quality assessment problem into two sub-tasks to promote the model to enhance understanding, focus attention and share feature information. Sub-task I classifies a DHH model into a specific distortion category. Sub-task II predicts the perceptual quality. The two sub-tasks share image features and together constitute the multi-task module. To analyze the performance of the proposed method, we select the DHHQA database \cite{zhang2022perceptual} for validation. The experimental results show that the proposed method achieves the best performance in the prediction accuracy of the quality scores, benefiting from the introduction of the multi-task module. 

\section{Proposed Method}

\noindent In this section, we describe our proposed NR DHHQA method in detail. The framework of the method is shown in Fig.~\ref{fig1}, which contains three modules: the projection module, feature extraction module and multi-task module. The projection module captures front-side projection of the DHH. Then the ViT is employed to extract the quality-aware features from the projection. Finally, the features are regressed into quality values and distortion categories by the multi-task module.
\subsection{Projection Module}
\noindent The projection module selects the front-side projection of the DHH because the front projection contains more facial features and details that most affect the perceptual quality. Given a DHH represented by a textured mesh, the geometry mesh structure $M$ can be defined as:
\begin{equation}
\setlength{\abovedisplayskip}{3pt}
\setlength{\belowdisplayskip}{3pt}
M \in \{ \{ v|v \in V\} ,\{ n|n \in N\} ,\{ e|e \in E\} \} ,
\end{equation}
where $V$, $N$, $E$ represent the set of the vertices, normal vectors and edges of the mesh respectively. The geometry center of the mesh can be calculated as:
\begin{equation}
\setlength{\abovedisplayskip}{3pt}
\setlength{\belowdisplayskip}{3pt}
\begin{array}{l}
{C_M} = \frac{1}{N}\sum\limits_{i = 1}^N {{v_i}}, \\
\end{array}
\end{equation}
where $N$ denotes the total number of vertices and ${C_M}$ denotes the center coordinates of DHH model. As shown in Fig.~\ref{fig:projection}, we use random region proposal to get input patches as $P$.
\subsection{Feature Extraction Module}
\noindent In view of the popularity and success of the Transformer applied to computer vision tasks, the proposed method uses ViT \cite{dosoVITskiy2020image} as the feature extraction backbone. The architecture of ViT we use is a Transformer Encoder structure consisting of $n$ blocks stacked on top of each other. Before entering the Encoder, ViT first takes the ${P}$ and further divides them into smaller patches. Next, we extract the feature maps of the ViT encoder output, and the process can be described as:
\begin{equation}
\setlength{\abovedisplayskip}{3pt}
\setlength{\belowdisplayskip}{3pt}
{F} = ViT({P}),
\end{equation}
where $ViT(\cdot)$ denotes the process of feature extraction using ViT as the backbone and ${F}$ represents the feature maps output from the ViT network. Next, the extracted feature maps are computed by average pooling to obtain a quality-aware embedding, and the process can be expressed as follow:
\begin{equation}
\setlength{\abovedisplayskip}{3pt}
\setlength{\belowdisplayskip}{3pt}
\overline{F} = Avg({F}),
\end{equation}
where $\overline{F}$ denotes the final extracted features of the feature module, and $Avg(\cdot)$ denotes the operation of average pooling.
\subsection{Multi-Task Module}
\noindent Specifically, the quality assessment problem is divided into two sub-tasks with the aim of enabling the method to perform the classification task while introducing additional quality information to improve the prediction accuracy of the quality scores. Sub-task I classifies a DHH model into a specific distortion category. Sub-task II predicts the perceptual quality. The two sub-tasks share image features and jointly optimize the network parameters to improve the performance of the visual perception task.\\
\indent After extracting features with perceptual quality significance, we use fully connected (FC) layers to aggregate them into quality scores for DHHs. For the classification task, we obtain the category probabilities by a two-layer FC layer consisting of 128 neurons and $N_d$ ($N_d$ denotes the number of distortion categories) neurons which represent the possibilities belonging to each category. For the regression task, the quality scores are derived from two FC layers consisting of 1024 neurons and 1 neuron respectively. Specifically, the process can be described as:
\begin{equation}
\setlength{\abovedisplayskip}{3pt}
\setlength{\belowdisplayskip}{3pt}
\begin{gathered}
  \widehat c = \alpha (\overline F ), \hfill \\
  \widehat q = \beta (\overline F ), \hfill \\ 
\end{gathered} 
\end{equation}
where $\widehat c$, $\widehat q$ are the distortion categories and quality scores predicted by the model respectively, $\alpha(\cdot)$, $\beta(\cdot)$ denote the regression of Sub-task I and Sub-task II respectively. In both tasks, we use the mean squared error (MSE) as the loss function. Thus, for the whole network, the loss function can be expressed as:
\begin{equation}
\setlength{\abovedisplayskip}{3pt}
\setlength{\belowdisplayskip}{3pt}
Loss = \frac{1}{{Ns}}\sum\limits_{i = 1}^{Ns} {(\lambda {{\left\| {{{\widehat c}_{{\text{i }}}} - {c_{{\text{i }}}}} \right\|}^2} + {{\left\| {{{\widehat q}_{{\text{i }}}} - {q_{{\text{i }}}}} \right\|}^2})},
\end{equation}
where $N_{\text s}$ is the number of samples in the mini-batch, ${c_{{\text{i }}}}$, ${q_{{\text{i }}}}$ denote the ground truth distortion categories and quality scores respectively, and $\lambda $ is a coefficient factor that is used to subjectively adjust the relative importance of the classification task and the regression task.


\section{Experiment Evaluation}
\subsection{Experiment Details.}
\noindent The proposed method is validated on DHHQA database consisting of 1,540 digital human head distortion projections. The database collects 55 scanned models of digital human heads and introduces seven distortions on the 3D models. Each distortion provides four different distortion levels and corresponding mean opinion scores (MOS). To test the generalization ability of the model, five-fold cross-validation strategy is employed. Specifically, we split the database into 5 folds and each fold contains 11 groups of DHHs. Then we select 4 folds for training and leave 1 fold for testing. Such process is repeated 5 times so that every fold has been used for testing.  The average performance of the 5 folds is recorded as the final results. It's worth mentioning that there is no content overlap between the training and testing sets. The selected ViT backbone which contains 12 Transformer Blocks is pre-trained on ImageNet-21k\cite{deng2009imagenet}. The Adam optimizer\cite{kingma2014adam} with an initial learning rate of 1e-5 is utilized. The default number of training epochs and batch size are set as 50 and 30 respectively. We split the dataset into five folds, and calculate the average of the above evaluation criteria as the final performance.

\subsection{Experiment Criteria.}
\noindent We compared the proposed method with ten NR IQA methods, including  BRISQUE\cite{mittal2012no}, CPBD\cite{narvekar2009no}, IL-NIQE\cite{zhang2015feature}, NFERM\cite{gu2014using}, NFSDM\cite{gu2013no}, NIQE\cite{mittal2012making}, DBCNN\cite{zhang2020blind}, StairIQA\cite{sun2021blind}. Among these ten NR IQA methods, CPBD, IL-NIQE, and NIQE do not require training. Besides, we also include the performance of some classic full-reference point cloud quality assessment (FR PCQA) methods such as p2point\cite{cignoni1998metro}, p2plane\cite{tian2017geometric} and psnr-yuv\cite{torlig2018novel} through converting the DHHs from textured meshes to point clouds. The default experimental setup is maintained for the rest of the compared methods.\\ 
\indent To compare the performance of these methods in the assessment of the visual perceptual quality of the digital human head, we use four commonly used evaluation criteria to evaluate different NR IQA methods, namely, Spearman rank order correlation coefficient (SRCC), Pearson linear correlation coefficient (PLCC), Kendall rank order correlation coefficient (KLCC), and root mean square error (RMSE). \\

\subsection{Performance Comparison with Other Methods}
\noindent The experimental results on the DHHQA database are exhibited in Table~\ref{tab1}. From Table~\ref{tab1}, we can see that the proposed method outperforms the other NR IQA metrics and leads by a significant margin (better than the second-place method StairIQA by 12.20\%). Firstly, this indicates that the proposed method is much more effective than the NR IQA based on handcrafted extracted features in assessing the quality of DHH models. It can be explained that most methods based on handcrafted extracted features are mainly designed for natural scene statistics (NSS). However, the rendered images are not necessarily subject to the prior distribution of NSS. Besides, in allusion to the fact that the proposed method outperforms some deep-learning based NR IQA methods, we speculate that the multi-task module can obtain more quality-aware information derived from the classification of distortion, which promotes the improvement of the performance. Finally, the classic FR PCQA methods are not effective for DHHQA task. It can be explained that these classic FR PCQA methods are not suitable for dealing with relatively more complicated models.
\subsection{Ablation Experiments}
\noindent In this section, to further validate the effectiveness of multi-task learning, we conduct the following ablation experiments. We set up two groups of models for comparison. One group is the baseline model based on ViT which only includes the Sub-task II, and the other group is the ViT model with multi-task module which can realize the Sub-task I and Sub-task II. The default experiment setup is maintained and the results of the two groups of models are shown in Table~\ref{tab3}. As can be seen from Table~\ref{tab3}, the multi-task module enables the model to obtain better performance than the baseline model in visual quality perception, indicating that the involvement of the Sub-task I can help improve the model's understanding of visual quality for DHHs.
\begin{table}\centering
\caption{Performance comparison of different methods. The best performance results are marked in \textbf{\textcolor{red}{RED}} and the second performance results are marked in \bf{\textcolor{blue}{BLUE}}.}
\label{tab1}
\setlength{\tabcolsep}{1mm}
\begin{tabular}{c|c|c|c|c|c}
\toprule
Type & Method &  SRCC$\uparrow$ & PLCC$\uparrow$ & KRCC$\uparrow$ & RMSE$\downarrow$\\
\hline
\multirow{3}{25pt}{\centering FR}
& MSE-p2ponit & 0.2891 & 0.29116 & 0.2359 & 21.0813\\
& MSE-p2plane & 0.2698 & 0.2961 & 0.2250 & 21.0502\\
& psnr-yuv & 0.1761 & 0.2272 & 0.1369 & 21.4299\\
\hdashline
\multirow{11}{25pt}{\centering NR}
& BRISQUE & 0.6008 & 0.5958 & 0.4214 & 17.3669\\
& CPBD & 0.2621 & 0.2599 & 0.1718 & 20.6701\\
& IL-NIQE & 0.5419 & 0.6232 & 0.3825 & 17.4119\\
& NFERM & 0.6414 & 0.6876 & 0.4610 & 15.4528\\
& NFSDM & 0.5921 & 0.6598 & 0.4214 & 16.1233\\
& NIQE & 0.4232 & 0.4410 &0.2784 & 18.9814\\
& DBCNN & 0.7575 & 0.8238 &0.5723 & 12.0087\\
& StairIQA & \bf{\textcolor{blue}{0.8052}} & \bf{\textcolor{blue}{0.8271}} & \bf{\textcolor{blue}{0.6123}} & \bf{\textcolor{blue}{11.8403}}\\
& Proposed & \bf{\textcolor{red}{0.9272}} & \bf{\textcolor{red}{0.9275}} & \bf{\textcolor{red}{0.7711}} & \bf{\textcolor{red}{7.4662}}\\
\bottomrule
\end{tabular}
\vspace{-0.6cm}
\end{table}

\begin{table}\centering
\caption{Performance of two models for ablation study. I indicates the Sub-task I and II indicates the Sub-task II. The best performance results are marked in \textbf{BOLD}.}\label{tab3}
\setlength{\tabcolsep}{1.9mm}
\begin{tabular}{c c|l|l|l|l|c}
\toprule
I & II  & SRCC$\uparrow$ & PLCC$\uparrow$ & KRCC$\uparrow$ & RMSE$\downarrow$ & ACC$\uparrow$\\
\hline
\ding{53} &  \checkmark & 0.9167 & 0.9173 & 0.7451 & 7.7209 & -\\
\checkmark &  \checkmark & \bf{0.9272} & \bf{0.9275} & \bf{0.7711} & \bf{7.4662} & \bf{0.9732}\\
\bottomrule
\end{tabular}
\vspace{-0.6cm}
\end{table}

\subsection{Performance Comparison of Different Backbones}
\noindent We compare the performance of three popular backbones, ResNet50\cite{he2016deep}, MobileNetV2\cite{sandler2018mobilenetv2}, and ViT\cite{dosoVITskiy2020image}. The performance of the three backbones is exhibited in Table~\ref{tab2}. From Table~\ref{tab2}, we can see that MobileNetV2 yields similar performance compared with ViT and conclude that firstly, our proposed method is still valid on a lightweight platform. Second, ViT achieves the best performance on the regression task of visual perception while maintaining about the same accuracy of distortion classification as ResNet50, further demonstrating the rationality of selecting ViT as the backbone.
\begin{table}\centering
\caption{Performance comparison of different backbones. MNV2 indicates the MobileNetV2. The best performance results are marked in \textbf{BOLD}.}\label{tab2}
\setlength{\tabcolsep}{1.6mm}
\begin{tabular}{c|l|l|l|l|l}
\toprule
Backbone &  SRCC$\uparrow$ & PLCC$\uparrow$ & KRCC$\uparrow$ & RMSE$\downarrow$ & ACC$\uparrow$\\
\hline
MNV2 & 0.9086 & 0.9043 & 0.7331 & 9.0336 & 0.9442\\
ResNet50 & 0.9210 & 0.9165 & 0.7519 & 8.4616 & 0.9669\\
ViT & \bf{0.9272} & \bf{0.9275} & \bf{0.7711} & \bf{7.4662} & \bf{0.9732}\\
\bottomrule
\end{tabular}
\vspace{-0.4cm}
\end{table}

\begin{figure}[t]
\begin{minipage}[b]{.49\linewidth}
  \centering
\centerline{\epsfig{figure=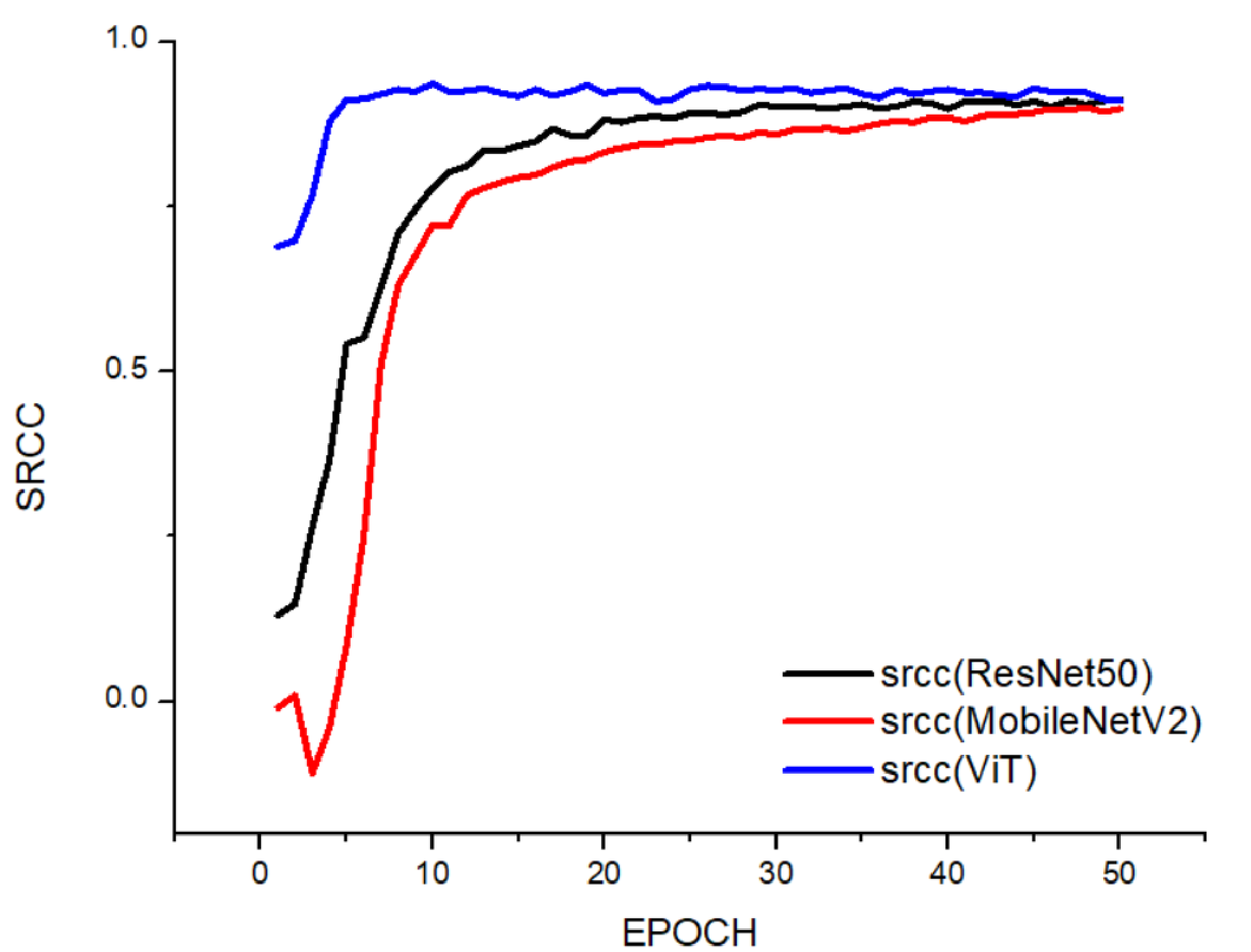,width=4.4cm}}
  \vspace{0.1cm}
  \centerline{(a) SRCC performance}\medskip
\end{minipage}
\begin{minipage}[b]{.49\linewidth}
  \centering
\centerline{\epsfig{figure=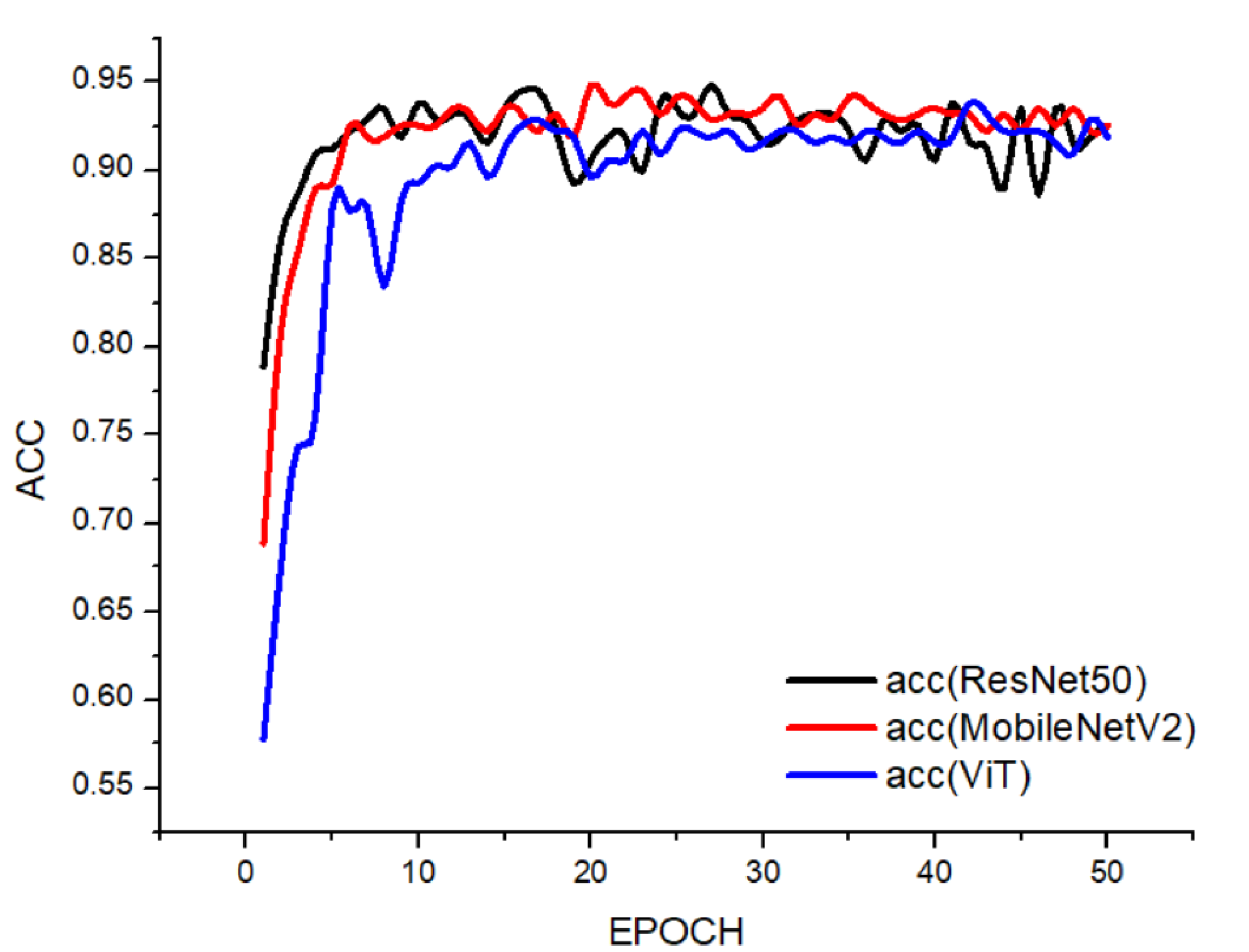,width=4.4cm}}
  \vspace{0.1cm}
  \centerline{(b) ACC performance}\medskip
\end{minipage}

\vspace{-0.5cm}
\caption{Different backbone training process.}
\vspace{-0.5cm}
\label{fig:performance}
\end{figure}
\indent To further discuss the performance of different backbones. We record the process of training the same fold time for three backbones and the result is shown in Fig.~\ref{fig:performance}. From Fig.~\ref{fig:performance}(a) we can conclude that our proposed method converges faster and better when dealing with the regression task of visual perception compared to ResNet50 and MobileNetV2. Moreover, according to Fig.~\ref{fig:performance}(b), it can be seen that ViT converges slower in the first few training rounds when dealing with the classification task, but with the increase of the num of epochs, it also eventually gets about the same or even better performance than ResNet50. In response to this result, we speculate that: CNN has a stronger feature extraction ability and is more likely to capture the difference between various distortions; while attention-based ViT is better at global visual quality perception from a broader view.
\section{Conclusion}
In this paper, we propose a novel Transformer-based NR 3DQA multi-task method for DHHQA. In order to reduce the computational cost of the method, we first record the front-side projection of the DHH model and crop the projection into patches. Then, we use ViT to perform feature extraction on the patches to obtain quality-awareness features. Specifically, we use the average pooling to process the output feature maps and then obtain the final quality-awareness embedding. Finally, the multi-task module is involved to regress quality-awareness embedding into quality scores as well as distortion classification respectively. The experimental results show that the proposed method significantly outperforms other methods on DHHQA and achieves competitive performance on the classification task, which validates the effectiveness of the proposed method for predicting the visual quality of DHHs.
\bibliographystyle{IEEEbib}
\bibliography{icme2022template}

\end{document}